%% file: main.tex
\documentclass[10pt, a4paper]{article}

\usepackage[final]{lrec2026} %

\usepackage{booktabs}
\usepackage{fontawesome5}

\title{Resource-Lean Lexicon Induction for German Dialects}

\name{Robert Litschko$^{1,2}$ \quad Barbara Plank$^{1,2}$ \quad Diego Frassinelli$^{1}$} 

\address{\textsuperscript{1}MaiNLP, Center for Information and Language Processing, LMU Munich, Germany \\
  \textsuperscript{2}Munich Center for Machine Learning (MCML), Munich, Germany\\
  \texttt{robert.litschko@lmu.de}}

\abstract{
Automatic induction of high-quality dictionaries is essential for building lexical resources, yet low-resource languages and dialects pose several challenges: limited access to annotators, high degree of spelling variations, and poor performance of large language models (LLMs). We empirically show that statistical models (random forests) trained on  string similarity features are surprisingly effective for inducing German dialect lexicons. They outperform LLMs, enable cross-dialect transfer, and offer a lightweight  data-driven alternative. We evaluate our models intrinsically on bilingual lexicon induction (BLI) and  extrinsically on dialect information retrieval (IR). On BLI, random forests outperform Mistral-123b while being more resource-lean. On dialect IR with BM25, using our dialect dictionaries for query expansion yields relative improvements of up to 28.9\% in nDCG@10 and 50.7\% in Recall@100. Motivated by the resource scarcity in dialects, we further investigate the extent to which models transfer across different German dialects, and their performance under varying amounts of training data. 
 \\ \newline  }

\begin{document}

\maketitleabstract

\section{Introduction}

The performance of current natural language processing (NLP) tools and large language models (LLMs) crucially depends on the language of use and how well it is represented in the (pre-)training corpus. High-resource languages like English and German benefit from robust machine translation (MT) and mature information retrieval (IR) systems. However, their performance has been shown to deteriorate when translating non-standard languages \citep{gupta2025endive,ziems-etal-2023-multi} or retrieving relevant information where entities appear in regional spelling variations and word choices \citep{litschko-etal-2025-cross,valentini2024messirve,chari2023effects}. This leads to a lexical dialect gap between majority languages and dialects. 

A major obstacle preventing the widespread adoption of NLP tools towards regional language varieties is the lack of resources and pretraining data, combined with the high variability due to the lack in standard orthography of many dialects. Multilingual language models cover only 1\% of the world's over 7,000 languages in existence \citep{wang2022expanding} and fail to accurately represent (i.e., tokenize) dialects at the input level \citep{munoz-ortiz-etal-2025-evaluating,srivastava-chiang-2025-calling,blaschke-etal-2023-manipulating}. Dictionaries offer a resource-lean alternative to capture dialect variation. For example in the context of retrieval, dictionaries have been traditionally used to translate queries to the document language \citep{ballesteros-croft-1996,adriani-rijsbergen-1999} and to induce cross-lingual embedding spaces \citep{litschko2018unsupervised}. In the context of translation, dictionaries have for example been used for data augmentation \citep{kalein-eta-2020,waldendorf2022improving,yin-etal-2024-lexmatcher}, or to improve the ability of LLMs to translate texts into low-resource languages and rare words \citep{ghazvininejad2023dictionary,lu-etal-2024-chain}. Compared to the (often unknown) language coverage of existing language models and translation systems, lexical resources exists for thousands of languages \citep{wang2022expanding}, with PanLex being the most prominent example \citep{baldwin-etal-2010-panlex,kamholz2014panlex}. For dictionaries to be useful in practical applications they need to have a high lexical coverage (high recall), and high translation quality (high precision) to minimize erroneous translations. Meeting the coverage requirement is a particularly challenging task in the case of building dialect variation dictionaries, where words are mapped to multiple spelling variations. Manually curated dialect dictionaries developed by linguists achieve the highest quality standards, but are often unavailable in machine-readable form. We therefore focus on inducing bilingual dictionaries in a data-driven and lean way based on real-world dialect usage. Our method assumes to have access to Wikipedia for the target dialect, but is otherwise generalizable. To test the effectiveness of our approach, in this paper we specifically focus on five German dialects, each of which has its own Wikipedia: Alemmanic (als), Bavarian (bar), Ripuarian (ksh), Rhine Franconian (pfl), and Low German (nds). 

Prior work on inducing German dialect dictionaries relied heavily on human supervision \citep{artemova2023low,litschko2025make-every-letter-count,litschko-etal-2025-cross}. \citet{artemova2023low} induce dictionaries based on word-alignments extracted from human-verified parallel sentences. This is not only costly and time-intensive due to the human supervision, but it also suffers from the same limitations as multilingual language models (i.e., most language varieties are not covered). Recent work tests LLMs for dictionary induction, but as shown in multiple studies \citep[i.e.,][]{li-etal-2023-bilingual,merx-etal-2024-generating,litschko2025make-every-letter-count}, building dictionaries with LLMs performs poorly in low-resource scenarios, especially for dialects. In contrast to these works, we show that statistical models trained on string similarity features are effective for building high-quality,  high-coverage dictionaries at a much cheaper computational cost than LLMs, while achieving better overall quality. This opens up the support for more inclusive technology for minority languages. We further ablate the model performance with respect to the amount of training data used in the BLI task to determine how much data is truly needed to build high-quality resources. 
We make the following contributions:
\begin{itemize}
    \item We create dialect variation dictionaries for five German dialects (§\ref{sec:dataset_evaluation}).
    
    \item We validate the quality of our statistical model intrinsically on the task of BLI (§\ref{ssec:bli}), showing that it outperforms Mistral-123b. 
    
    \item We evaluate our induced dictionaries extrinsically on the task of cross-dialect information retrieval (§\ref{ssec:cdir}), showing consistent improvements across all dialects. 
\end{itemize}

We make our code and resources available for future uptake.\footnote{\href{https://github.com/mainlp/dialect-lexicon-induction}{https://github.com/mainlp/dialect-lexicon-induction}}

\section{Methodology}\label{sec:method}

\subsection{Classifying Dialect Variation}
We frame the task of bilingual induction as a word-pair classification task. The starting point for this is a vocabulary in German $l_{de}$ (majority language) and in a regional language variety $l_{dial}$ (dialect). The goal is to match terms in $l_{de}$ to one or more terms in $l_{dial}$. Since the quadratic combination of all possible matches is too large to explore in practice, we use the \textsc{DiaLemma} annotation framework \citep{litschko2025make-every-letter-count} to obtain for each term up to $k=10$ lexical nearest neighbors, using Levenshtein distance \citep{levenshtein-1966-binary}. This allows us to pre-filter promising candidate dialect words prior to scoring word-pairs based on more sophisticated string similarity features.

\subsection{String Similarity Features}\label{sec:string_sim}
In this section, we describe the string similarity features used for our models, most of which have been used in \citet{inkpen2005automatic} in the context of classifying word pairs as cognates or false friends. The similarity measures are computed on pairs consisting of a German lemma $x$ and a candidate dialect term $y$, and can be broadly grouped into set-based and sequence-based measures. 

Set-based measures first transform the dialect term $y$ and German term $x$ into sets of ngrams. The first measure has been proposed by \citet{adamson1974use} and corresponds to the Dice-Sørensen coefficient computed on the shared character ngrams:
\begin{equation}
    \textsc{Dice}(x, y) = \frac{2\cdot|ngrams(x) \cap ngrams(y)|}{|ngrams(x)+ngrams(y)|}
\end{equation}
Following \citet{inkpen2005automatic}, we apply \textsc{Dice} on the set of bigrams, trigrams, and so-called ``extended trigrams'' consisting of trigrams minus their middle character \citep[\textsc{XDice};][]{brew1996word}. \textsc{XXDice} extends \textsc{XDice} by incorporating positional information, each overlapping token is weighted by 
\begin{equation}
    \frac{1}{1+(pos(a)-pos(b))^2}\raisebox{0.35ex}{,}
\end{equation}
where $pos(a)$ and $pos(b)$ correspond to the positions of the shared token in $x$ and $y$. If a shared token appears multiple times we take the last occurrence. This scaling factor reduces the influence of matching ngrams that appear in different positions. 

Sequence-based string similarity compare how well pairs of strings are aligned. The first feature simply measures the length the common prefix found in $x$ and $y$ (\textsc{Prefix}). The longest common subsequence ratio \citep[\textsc{LCSR};][]{melamed1999bitext} relaxes the constraint that characters need to be adjacent, it counts the number of characters that appear in the same order divided by the length of the longer string. Following \citet{inkpen2005automatic}, we also use the extended version of \textsc{LCSR}, which works on sequences of character bigrams (\textsc{Bi-Sim}) and trigrams (\textsc{Tri-Sim}) instead of sequences of individual characters \citep{kondrak-dorr-2004-identification}. In addition to sequence-based string similarity, we use the edit distance between $x$ and $y$, normalized by the length of the longer sequence (\textsc{NED}) \citep{wagner1974string}. We also compute the NED on bigram (\textsc{Bi-Dist}) and trigram sequences (\textsc{Tri-Dist}) \citep{inkpen2005automatic}. 
Our final string similarity feature is based on phonetic similarity. We first encode both strings using the cologne phonetics algorithm \citep{postel1969kolner} to transform both strings to their phonetic code, which encodes their pronunciations, and then compute the length-normalized edit distance between the two codes. This allows us  match of similar-sounding words despite spelling differences. We deliberately use only string similarity features to evaluate how well pairs of German lemmas and dialect terms can be matched based solely on their surface form. The performance of our models can likely be improved by incorporating additional information such as term frequencies, part of speech information, or semantic features.

\subsection{Statistical Model}
In this work, we use Random Forest \citep{breiman2001random} implemented in the scikit-learn library \citep{pedregosa2011scikit}. We resort to the default values, where each random forest consists of 100 decision trees, using Gini impurity as a splitting criterion during training. In contrast to \citet{litschko2025make-every-letter-count}, which applies logistic regression as a linear classification model, random forest allows for fitting non-linear decision boundaries. For each dialect and dataset (see §\ref{sec:dataset_evaluation}), we train our model on 80\% of the data, leaving 20\% for testing. All reported results correspond to the average over repeated experiments with three different random seeds. A core advantage of statistical models is that they are much less resource-demanding and faster than LLMs, but also more effective in identifying dialect variations (§\ref{sec:results}).

\subsection{Datasets and Evaluation}\label{sec:dataset_evaluation}
\paragraph{Bilingual Lexicon Induction (BLI).} We evaluate our model intrinsically by measuring its performance on inducing bilingual dictionaries. Following prior work \citep{heyman-etal-2017-bilingual,irvine-callison-burch-2017-comprehensive}, we treat BLI as a classification task. Models are presented with a German word and a dialect candidate and must predict whether they represent translations of one another. Resorting to classification measures serves as a proxy to measure the dictionary quality in terms of its coverage (recall) and proportion of false entries (precision). 

We evaluate on two recently published dialect variation dictionaries: \textsc{DiaLemma} \citep{litschko2025make-every-letter-count} and \textsc{WikiDIR} \citep{litschko-etal-2025-cross}. \textsc{DiaLemma} consists of 100k Bavarian word pairs in different word classes, while \textsc{WikiDIR} covers entities in five German dialects: Alemannic (als), Bavarian (bar), Low German (nds), Rhine Franconian (pfl), and Ripuarian (ksh). Both datasets are human-annotated, \textsc{DiaLemma} is uses three classes to indicate if a term is a dialect translation, an inflected variant, or unrelated, while \textsc{WikiDIR} adopts a binary label scheme. The datasets also differ in how candidate words were sourced: \textsc{DiaLemma} dictionaries are based on lexical nearest neighbors, while in \textsc{WikiDIR} they are derived from  inter-language and inter-article links on Wikipedia. 

\paragraph{Dialect retrieval.} We evaluate our model extrinsically on the task of cross-dialect information retrieval. %
We specifically use the \textit{analysis split}, where relevant documents contain query keywords in different spelling variations. For retrieval, we use the BM25 model \citep{robertson2009probabilistic} implemented in the Pyserini library \citep{lin2021pyserini}. Lexical retrieval models extract relevance signals from exact keyword matches and are ineffective when documents contain dialect variations of those keywords, since most dialects lack lexical normalization tools, such as stemmers and lemmatizers. We compare the performance of BM25 with the original queries against expanded queries, where we look up and append dialect spelling variations of query keywords. We measure the extent to which query expansion (QE) improves the results in terms of nDCG@10 and Recall@100. 

For this experiment, we use the annotation framework proposed in \citet{litschko2025make-every-letter-count} for our five German dialects. For each dialect we 1) collect the 100K most frequent German lemmas, 2) find for each lemma its ten nearest lexical neighbors, and 3) classify word pairs whether they correspond to translations. %
The classification model used for step 3) is trained on the full \textsc{DiaLemma} dictionary (5.2K lemmas). We construct our dialect dictionaries by including all word pairs that the model identified as translation equivalents. The resulting dataset statistics are shown in Table~\ref{tab:multilemma_dataset}.

{
\setlength{\tabcolsep}{10pt}
\begin{table}[t!]
    \centering
\begin{tabular}{lrrr}
\toprule
\textbf{Dialect} & \textbf{Lemmas} & \textbf{Variants} & \textbf{V/L} \\
\midrule
\textbf{als} & 38,129 & 88,114 & 2.31 \\
\textbf{bar} & 27,598 & 51,392 & 1.86 \\
\textbf{ksh} & 6,889 & 9,384 & 1.36 \\
\textbf{pfl} & 9,127 & 13,050 & 1.43 \\
\textbf{nds} & 21,974 & 39,547 & 1.80 \\
\bottomrule
\end{tabular}
    \caption{Dataset statistics of our induced dictionaries. We show the number of lemmas for which we found at least one dialect variant, the total number of dialect terms, and the average per lemma.}
    \label{tab:multilemma_dataset}
\end{table}
}

\input{results/statistical_vs_llm}

\section{Results and Discussion}\label{sec:results}

\input{results/transfer}

\subsection{Bilingual Lexicon Induction}\label{ssec:bli}
Table~\ref{tab:cross-dialect-f1} reports the $F_1$ scores for BLI across the five dialects in \textsc{WikiDIR}. The last row shows the results of a multi-source model that is trained on the concatenation of all training splits. Overall, the model achieves a good performance ($F_1 = 0.75 \pm 0.07$) when trained and tested on the same dialect (i.e., the values on the diagonal). This confirms that the model can capture dialect-specific regularity in an effective way. When transferring across dialects, we observe varying results.  When training on a single dialect, Alemannic shows the best results across all dialects (als row, $F_1 = 0.75 \pm 0.05$). In contrast, models trained on Low German data show a lower transferability (nds row, $F_1 = 0.63 \pm 0.03$). Across the board, combining the training data of all languages (ALL) achieves the highest $F_1$ scores across $F_1 = 0.76 \pm 0.05$), indicating that the diversity and volume of training data positively impact cross-dialect transfer. 

Tables~\ref{tab:cross-dialect-precision}~and~\ref{tab:cross-dialect-recall} show that Low German (nds) and Ripuarian (ksh) stand out as the best source languages, yielding the highest precision and recall values across all dialects. This suggests that the training data of Low German has a more consistent or less varied feature distribution, while the training data of Ripuarian might capture a more diverse set of features. Further research is needed to investigate the factors that influence cross-dialectal transfer at the lexical level. 

\input{results/training_data_size}

\subsection{Effect of Training Size}\label{ssec:bli-2}
Figure~\ref{fig:perf_vs_size} shows the effect of training size on model performance when trained on different portions of the training data and evaluated on the same 20\% test split. We use word pairs from the \textsc{DiaLemma} dictionary, where full training and test datasets contain 80k and 20k training instances. We find that using only 10\% of the training data (8k word pairs) yields an average $F_1$ score of 0.52, and training on 40\% of the data  ($F_1=0.56$) outperforms Mistral (see Table~\ref{tab:bar-F1-comparison}). A further increase in training size leads to a modest improvement of 5\% ($F_1=0.59$). This indicates that even a relatively small training set is sufficient to obtain a strong performance, which is promising for dialects where data is scarce.

\input{results/retrieval}

\subsection{Cross-Dialect Information Retrieval}\label{ssec:cdir}
Table~\ref{tab:ndcg-recall} reports the results of the cross-dialect retrieval experiments using BM25 as baseline and a query-expanded (QE) variant across five dialects. As indicated by the positive deltas, query expansion consistently improves the retrieval results of BM25. On average, we observe an improvement (relative improvement) of +0.05 nDCG@10 (+14.71\%) and +0.08 Recall@100 (+25.1\%). This improvement is particularly evident for Rhine Franconian (pfl), which is the dialect with the lowest overall number of queries and the smallest document corpus. However, despite its smaller scale, BM25 (without QE) achieves the lowest performance on Rhine Franconian. In terms of coverage, we observe that approximately 51\% of all German queries included at least one keyword for which dialect spelling variations are available in our dictionaries.

\section{Conclusion}
In this work, we show that statistical models trained on elaborate string similarity features are not only more resource-lean, both in terms of (pre-)training requirements and at inference time, but also more effective. We contribute dialect variation dictionaries for five German dialects, covering more dialects than \textsc{DiaLemma} (5 vs.\ 1), and substantially more lemmas than \textsc{WikiDIR} (103,717 vs.\ 6,257). In our extrinsic evaluation, we show that query expansion with our dictionaries consistently improves the dialect retrieval performance with BM25.

\section{Limitations}
In this work, we focus exclusively on surface-level features and do not incorporate (contextual) semantic similarity. While this has the advantage of being resource-efficient, it does not account for ambiguous terms. However, this affects only a minority of the cases. In future work, we plan to jointly model orthographic and semantic features.

\section{Ethical Considerations}
We see no ethical issues related to this work. All experiments were conducted with publicly available data and open-source software, and our code and linguistic resources are openly available on GitHub.

\paragraph{Use of AI Assistants} The authors acknowledge the use of ChatGPT for correcting grammatical errors and enhancing the coherence of the final manuscript.

\section{Bibliographical References}\label{sec:reference}

\bibliographystyle{lrec2026-natbib}
\bibliography{references}

\end{document}

%% file: results/statistical_vs_llm.tex
\begin{table}[t!]
\centering
\begin{tabular}{lccc}
\toprule
\textbf{Model} & \textbf{P} & \textbf{R} & \textbf{F1} \\ 
\midrule
\textbf{Random} & 0.112 & 0.341 & 0.169 \\ 
\textbf{Mistral-123b} & 0.443 & \textbf{0.743} & 0.555 \\
\textbf{Random Forest} & \textbf{0.646} & 0.534 & \textbf{0.585} \\
\bottomrule
\end{tabular}
\caption{Comparison between our Random Forest model against the \citet{litschko2025make-every-letter-count} best large language model (Mistral-123b) and a random baseline on Bavarian.}
\label{tab:bar-F1-comparison}
\end{table}

%% file: results/transfer.tex
\begin{table}[t!]
\centering
\begin{tabular}{lccccc}
\toprule
\textbf{train / test} & \textbf{ksh} & \textbf{nds} & \textbf{als} & \textbf{pfl} & \textbf{bar} \\ 
\midrule
\textbf{ksh} & \textbf{0.79} & 0.62 & 0.77 & \textbf{0.82} & 0.67 \\
\textbf{nds} & 0.60 & 0.68 & 0.63 & 0.61 & 0.63 \\
\textbf{als} & 0.75 & \textbf{0.71} & \textbf{0.80} & 0.81 & \textbf{0.70} \\
\textbf{pfl} & \textbf{0.79} & 0.61 & 0.76 & \textbf{0.82} & 0.67 \\
\textbf{bar} & 0.72 & 0.67 & 0.72 & 0.70 & 0.68 \\ \midrule 
\textbf{ALL} & 0.77 & \textbf{0.71} & 0.80 & 0.81 & \textbf{0.70} \\

\bottomrule
\end{tabular}
\caption{BLI - Cross-dialect $\mathbf{F_1}$ scores, with models trained on a specific source dialect and tested on all five target dialects. The highest score for each dialect is shown in bold.}
\label{tab:cross-dialect-f1}
\end{table}

\begin{table}[t!]
\centering
\begin{tabular}{lccccc}
\toprule
\textbf{train / test} & \textbf{ksh} & \textbf{nds} & \textbf{als} & \textbf{pfl} & \textbf{bar} \\ 
\midrule 
\textbf{ksh} & 0.76 & 0.49 & 0.67 & 0.77 & 0.57 \\
\textbf{nds} & \textbf{0.93} & \textbf{0.77} & \textbf{0.80} & \textbf{0.94} & \textbf{0.75} \\
\textbf{als} & 0.84 & 0.66 & 0.78 & 0.87 & 0.67 \\
\textbf{pfl} & 0.78 & 0.49 & 0.68 & 0.80 & 0.58 \\
\textbf{bar} & 0.83 & 0.66 & 0.78 & 0.84 & 0.70 \\ \midrule 
\textbf{ALL} & 0.85 & 0.67 & 0.79 & 0.90 & 0.67 \\
\bottomrule
\end{tabular}
\caption{BLI - Cross-dialect \textbf{precision} scores, with models trained on a specific source dialect and tested on all five dialects. The highest score for each dialect is shown in bold.}
\label{tab:cross-dialect-precision}
\end{table}

\begin{table}[t!]
\centering
\begin{tabular}{lccccc}
\toprule
\textbf{train / test} & \textbf{ksh} & \textbf{nds} & \textbf{als} & \textbf{pfl} & \textbf{bar} \\ 
\midrule 
\textbf{ksh} & \textbf{0.81} & \textbf{0.85} & \textbf{0.89} & \textbf{0.87} & \textbf{0.82} \\
\textbf{nds} & 0.45 & 0.60 & 0.53 & 0.45 & 0.54 \\
\textbf{als} & 0.67 & 0.77 & 0.82 & 0.76 & 0.75 \\
\textbf{pfl} & 0.79 & 0.80 & 0.86 & 0.86 & 0.79 \\
\textbf{bar} & 0.64 & 0.69 & 0.67 & 0.59 & 0.66 \\ \midrule  
\textbf{ALL} & 0.71 & 0.76 & 0.81 & 0.74 & 0.73 \\
\bottomrule
\end{tabular}
\caption{BLI - Cross-dialect \textbf{recall} scores, with models trained on a specific source dialect and tested on all five target dialects. The highest score for each dialect is shown in bold.}
\label{tab:cross-dialect-recall}
\end{table}

%% file: results/training_data_size.tex
\begin{figure}[t!]
    \centering
    \includegraphics[width=1\linewidth]
    {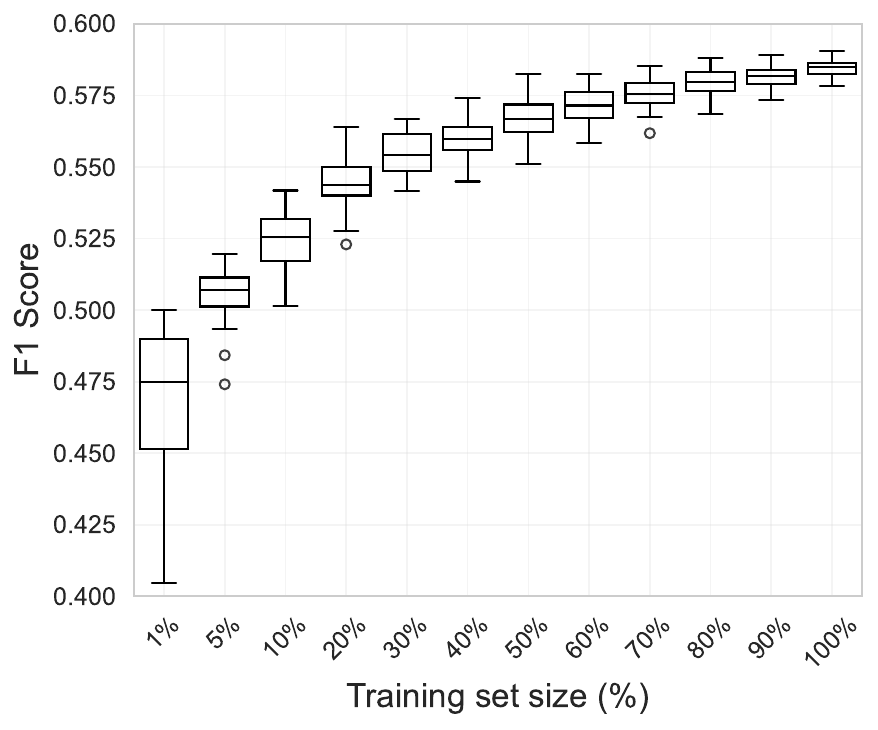}
    \caption{Effect of training set size on F1 scores for \textsc{DiaLemma} word pairs. We repeated our experiments with 40 different random seeds and evaluated on a fixed 20\% test split.}
    \label{fig:perf_vs_size}
\end{figure}

%% file: results/retrieval.tex
\begin{table*}[t!]
\centering
\resizebox{\textwidth}{!}{
\begin{tabular}{lcccr|cccr|rrr}
\toprule
\multicolumn{1}{c}{} & \multicolumn{4}{c|}{\textbf{nDCG@10}} & \multicolumn{4}{c|}{\textbf{Recall@100}} & \multicolumn{3}{c}{\textbf{Statistics}} \\ 
\textbf{} & \textbf{BM25} & \textbf{QE} & \textbf{$\Delta$} & \textbf{$\Delta$\%} & \textbf{BM25} & \textbf{QE} & \textbf{$\Delta$} & \textbf{$\Delta$\%} & \textbf{n\_aug} & \textbf{n\_query} & \textbf{\% n\_aug} \\ 
\midrule
\textbf{ksh}  & 0.33 & 0.38 & 0.06 & 18.2\% & 0.31 & 0.36 & 0.06 & 18.0\% &   135 &   210 & 64.3\% \\
\textbf{nds}  & 0.35 & 0.39 & 0.04 &  12.1\% & 0.28 & 0.34 & 0.06 & 20.0\% &   270 &   470 & 57.4\% \\
\textbf{als}  & 0.36 & 0.39 & 0.03 &  8.7\% & 0.34 & 0.43 & 0.09 & 27.5\% &  2021 &  4639 & 43.6\% \\
\textbf{pfl}  & 0.28 & 0.36 & 0.08 & 28.9\% & 0.21 & 0.31 & 0.11 & 50.7\% &   76 &   157 & 48.4\% \\
\textbf{bar}  & 0.45 & 0.48 & 0.03 &  5.7\% & 0.41 & 0.49 & 0.08 & 18.9\% &   298 &   718 & 41.5\% \\
\midrule 
\textbf{ALL}  & 0.35 & 0.40 & 0.05 & 14.71\% & 0.33 & 0.39 & 0.08 & 25.1\% &   560 &  1239 & 51.0\% \\
\bottomrule
\end{tabular}
}
\caption{Cross-Dialect Information Retrieval - We show for each dialect the result of applying BM25 on the original queries (BM25) and after query expansion (QE), as well as the absolute ($\Delta$) and relative differences ($\Delta \%$) in retrieval performance. \textbf{n\_aug} and \textbf{\%n\_aug} refer to the absolute and relative number of augmented queries. \textbf{n\_query} denotes the total number of queries.}
\label{tab:ndcg-recall}
\end{table*}